\documentclass[10pt,twocolumn,letterpaper]{article}

\usepackage{cvpr}
\usepackage{times}
\usepackage{epsfig}
\usepackage{graphicx}
\usepackage{amsmath}
\usepackage{amssymb}
\usepackage{color}
\usepackage{multirow}

\usepackage[pagebackref=true,breaklinks=true,letterpaper=true,colorlinks,bookmarks=false]{hyperref}

\cvprfinalcopy 

\def\cvprPaperID{3580} 

\ifcvprfinal\fi
\begin{document}

\title{AON: Towards Arbitrarily-Oriented Text Recognition}


\author{Zhanzhan Cheng\textsuperscript{1}\qquad\qquad
Yangliu Xu\textsuperscript{2}\qquad\qquad
Fan Bai\textsuperscript{3}\qquad\qquad  Yi Niu\textsuperscript{1} \\
Shiliang Pu\textsuperscript{1}\qquad\qquad\qquad
Shuigeng Zhou\textsuperscript{3}\thanks{Corresponding author.}\\
\textsuperscript{1}Hikvision Research Institute, China;\textsuperscript{2}Tongji University, Shanghai, China;\\
\textsuperscript{3}Shanghai Key Lab of Intelligent Information Processing, and School of \\Computer Science, Fudan University, Shanghai, China;\\
{\tt\small \{chengzhanzhan;xuyangliu;niuyi;pushiliang\}@hikvision.com;}\\
{\tt\small 1993sallyxyl@tongji.edu.cn;\{fbai17;sgzhou\}@fudan.edu.cn}
}

\maketitle

\begin{abstract}
  Recognizing text from natural images is a hot research topic in computer vision due to its various applications.
  Despite the enduring research of several decades on optical character recognition (OCR), recognizing texts from natural images is still a challenging task.
  This is because scene texts are often in irregular (e.g. curved, arbitrarily-oriented or seriously distorted) arrangements, which have not yet been well addressed in the literature.
  Existing methods on text recognition mainly work with regular (horizontal and frontal) texts and cannot be trivially generalized to handle irregular texts.
  In this paper, we develop the arbitrary orientation network (AON) to directly capture the deep features of irregular texts, which are combined into an attention-based decoder to generate character sequence. The whole network can be trained end-to-end by using only images and word-level annotations.
  Extensive experiments on various benchmarks, including the CUTE80, SVT-Perspective, IIIT5k, SVT and ICDAR datasets, show that the proposed AON-based method achieves the-state-of-the-art performance in irregular datasets, and is comparable to major existing methods in regular datasets.
\end{abstract}

\section{Introduction}\label{introduction}
Scene text recognition has attracted much research interest of the computer vision community~\cite{cheng2017focus,jaderberg2014synthetic,lee2016recursive,neumann2012real,shi2016end,Yang2017attention} because of its various applications such as road sign recognition and navigation reading for advanced driver assistant system (ADAS).
Though Optical Character Recognition (OCR) has been extensively studied for several decades, recognizing texts from natural images is still a challenging task due to complicated environments (e.g. uneven lighting, blurring, perspective distortion and orientation).

In the past years, there have been many works to solve scene text recognition~\cite{cheng2017focus,lee2016recursive,shi2016end,Yang2017attention}.
Although these approaches have shown promising results, most of them can effectively handle only regular texts that are often tightly-bounded, horizontal and frontal.
\begin{figure}[!thb]
    \begin{center}
    \includegraphics[width=0.45\textwidth]{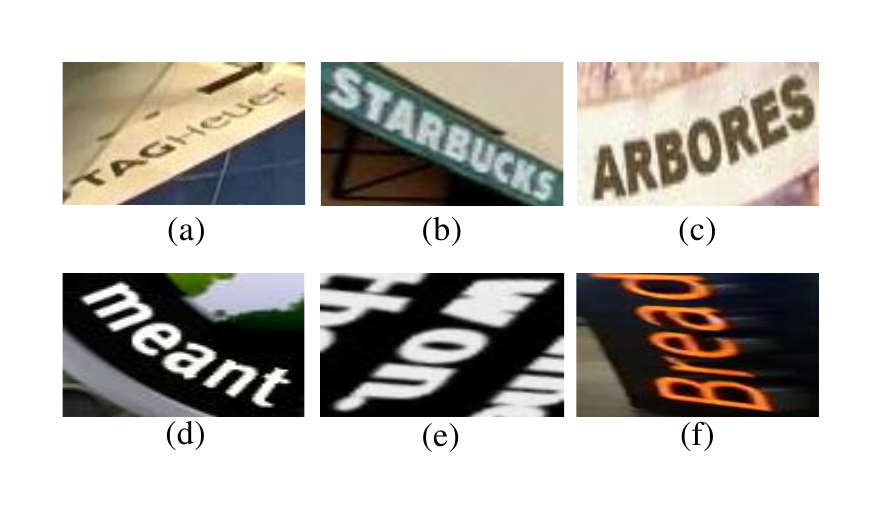}
    \end{center}
    \vspace{-2.3em}
    \caption{Examples of irregular (slant/perspective, curved and oriented etc.) texts in natural images. Subfigures (a) - (b), (c) - (d) and (e) - (f) are slant/perspective, curved and oriented images respectively.}
    \label{fig:hard}
\end{figure}
However, in real-world applications, many scene texts are in irregular arrangements (e.g. arbitrarily-oriented, curved, slant and perspective etc.) as shown in Fig.~\ref{fig:hard}, so most existing methods cannot be widely applied in practice.

Recently, there are two related works aiming at  irregular texts: the spatial transformer network (STN)~\cite{Jaderberg2015} - based method by~\cite{shi2016robust} and the attention-based method with fully convolutional network (FCN) \cite{Long2015fcn} by~\cite{Yang2017attention}.
Shi \etal \cite{shi2016robust} attempted to first rectify irregular (e.g. curved or perspectively distorted) texts to approximately regular texts, then recognized the rectified images with an attention-based sequence recognition network.
However, in complicated (e.g. arbitrarily-oriented or serious curved) natural scenes, it is hard to optimize the STN-based method without human-labeled geometric ground truth. Besides, training STN needs sophisticated skills. For example, the thin-plate-spline (TPS)~\cite{Bookstein1989}-based STN~\cite{shi2016robust} should be given some initialization pattern for the fiducial points, and is not quite effective for arbitrarily-oriented scene texts.
Yang \etal \cite{Yang2017attention} introduced an auxiliary dense character detection task for encouraging the learning of visual representations with a fully convolutional network. Though the method showed better performance on irregular texts, it was carried out with an exhausting multi-task learning (MTL) strategy and relied on character-level bounding box annotations.
Note that, though the attention-based model has the potential to perform 2D feature selection~\cite{xu2016show}, we found in experiments that directly training attention-based model on irregular texts is difficult due to irregular character placements. This situation motivates us to explore new and more effective methods to recognize irregular scene texts.
\begin{figure}[!thbp]
\label{fig:motivation}
    \begin{center}
    \includegraphics[width=0.45\textwidth]{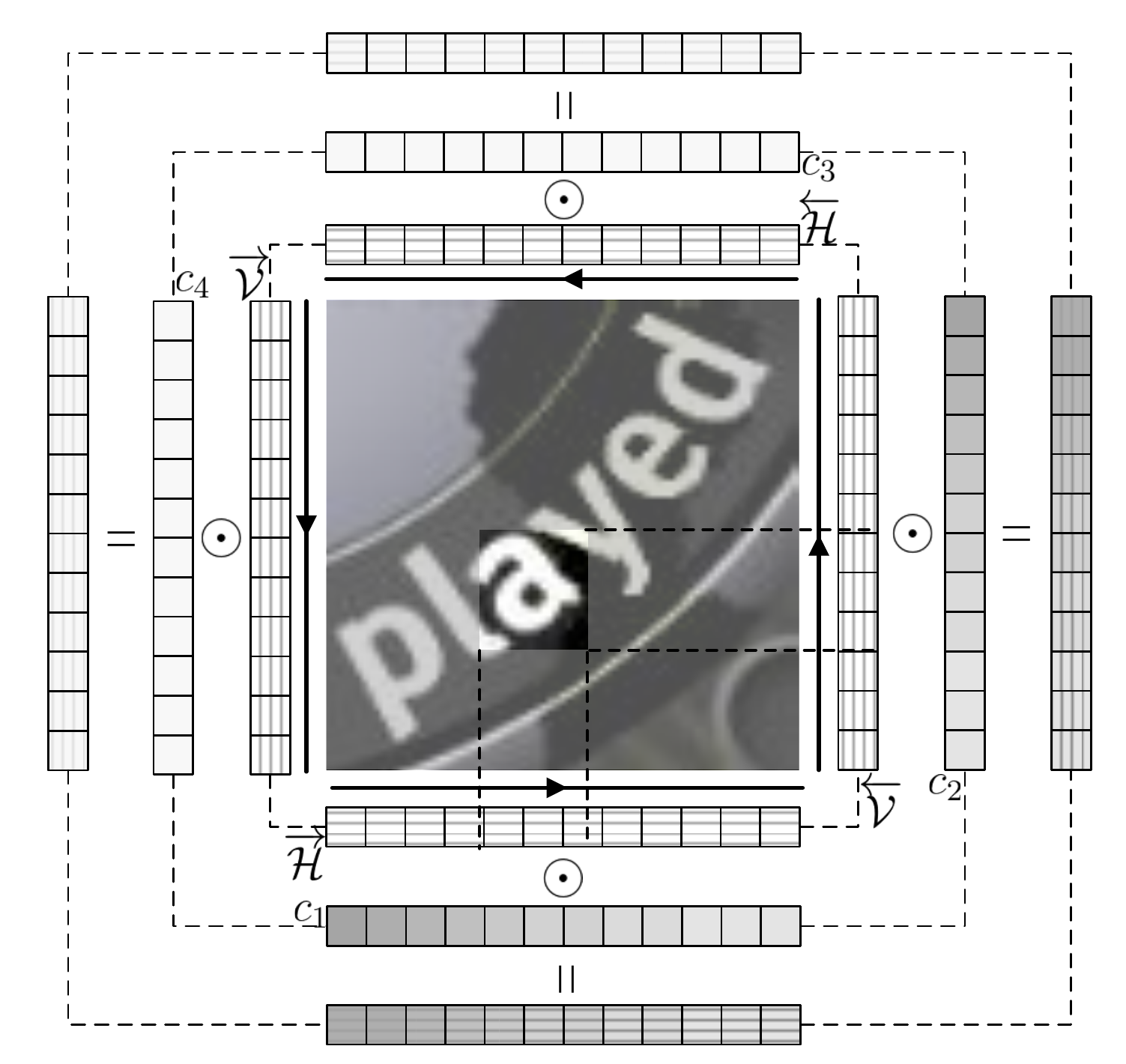}
    \end{center}
    \vspace{-0.5em}
    \caption{Illustration of character visual representation in four directions: $\protect\overrightarrow{{\mathcal{H}}}$: $left \rightarrow right$, $\protect\overleftarrow{{\mathcal{H}}}$:$right \rightarrow left$, $\protect\overrightarrow{{\mathcal{V}}}$:$top \rightarrow bottom$ and $\protect\overleftarrow{{\mathcal{V}}}$:$bottom \rightarrow top$ and four character placement clues $c_1, c_2, c_3$ and $c_4$. Here, there are three squares connected with dashed lines. The innermost square represents the four 1D sequences of features, each comes along with an arrowed line. The middle square refers to the placement clues used for weighting the corresponding sequences of features. The outermost square stands for the weighted feature sequence by conducting Hadamard product $\odot$ with character placement clues and horizontal/vertical features. For the character `a' in the image, we can represent it by the four weighted sequences of features.}
\end{figure}

From the above analysis, we can see that most existing methods directly encode a text image as a 1D sequence of features and then decode them to the predicted text, which implies that any text in an image is treated in the same direction such as \emph{from left to right} by default.
However, this is not true in the wild.
After carefully analyzing the typical character placement styles of natural text images, we suggest that the visual representation of an arbitrarily-oriented character in a 2D image can be described in four directions: $left \rightarrow right$, $right \rightarrow left$, $top \rightarrow bottom$ and $bottom \rightarrow top$.
Concretely, we can encode the input image to four feature sequences of four directions: \emph{horizontal features} ($\overrightarrow{{\mathcal{H}}}$), \emph{reversed horizontal features} ($\overleftarrow{{\mathcal{H}}}$), \emph{vertical features} ($\overrightarrow{{\mathcal{V}}}$) and \emph{reversed vertical features} ($\overleftarrow{{\mathcal{V}}}$), as shown in Fig.~\ref{fig:motivation}, and the length of each sequence is equal.
The horizontal/vertical features can be extracted by downsampling the height/width of feature maps to 1.
In order to represent an arbitrarily-oriented character, a weighting mechanism can be used to combine the four feature sequences of different directions. We call the weights \emph{character placement clues}, which are denoted as $c_1, c_2, c_3$ and $c_4$ in Fig.~\ref{fig:motivation}.
The character placement clues can be learned from the input images with a convolutional-based network, which guides to effectively integrate the four sequences of features, and then a filter gate (FG) generates the integrated feature sequence as the character's visual representation.
Therefore, an arbitrarily-oriented character in a 2D image can be represented as the combination of horizontal and vertical features by conducting the Hadamard product with the sequences of features and the corresponding placement clues.
In Fig.~\ref{fig:motivation}, $c_1$ and $c_2$ play the dominant role in determining the visual representation of character `a'.
In this paper, we call the four-direction feature extraction network and the clues extraction network \emph{arbitrary orientation network} (AON), which means that it can effectively handle arbitrarily-oriented texts.

In this paper, we develop a novel method for robustly recognizing both regular and irregular natural texts by employing the proposed \emph{arbitrary orientation network} (AON).

Major contributions of this paper are as follows:
\begin{enumerate}
\item We propose the arbitrary orientation network (AON) to extract scene text features in four directions and the character placement clues.
\item We design a filter gate (FG) for fusing four-direction features with the learned placement clues. That is, FG is responsible for generating the integrated feature sequence.
\item We integrate AON, FG and an attention-based decoder into the character recognition framework. The whole network can be directly trained end-to-end without any character-level bounding box annotations.
\item We conduct extensive experiments on several public irregular and regular text benchmarks, which show that our method obtains state-of-the-art performance in irregular benchmarks, and is comparable to major existing methods in regular benchmarks. 
\end{enumerate}

\section{Related works}
\label{related_works}
In recent years, several methods have been proposed for scene text recognition.
For the general information of text recognition, readers can refer to Ye and Doermann's recent survey~\cite{ye2015text}.
Basically, there are two types of scene text recognition approaches: bottom-up and top-down.

Traditional methods mostly follow the \emph{bottom-up} pipeline: first extracting low-level features for individual character detection and recognition one by one, then integrating these characters into words based on a set of heuristic rules or a language model.
For example,~\cite{neumann2012real} defined a set of handcrafted features such as aspect ratio, hole area ratio etc. to train a Support Vector Machine (SVM) classifier.
\cite{wang2011end,wang2010word} first fetched each character in the cropped word image by sliding window, then recognized it with a character classifier trained by the extracted HOG descriptors~\cite{yao2014strokelets}.
However, the performance of these methods is limited due to the low representation capability of handcrafted features.
With the advancement of neural-network-based methods, many researchers developed deep neural architectures and achieved better results.
\cite{bissacco2013photoocr} adopted a fully connected network of 5 hidden layers for character feature representation, then used an n-gram language model to recognize characters.
\cite{wang2012end} developed a CNN-based feature extraction framework for character recognition, and applied a non-maximum suppression method for final word predictions.
\cite{jaderberg2014deep} also proposed a CNN-based method with structured output layer for unconstrained recognition.
These above methods require the segmentation of each character, which can be very challenging because of the complicated background clutter and the inadequate distance between consecutive characters. Besides, segmentation annotations require additional resource consuming.
\begin{figure}[!htbp]
\vspace{-1em}
    \begin{center}
    \includegraphics[width=0.49\textwidth]{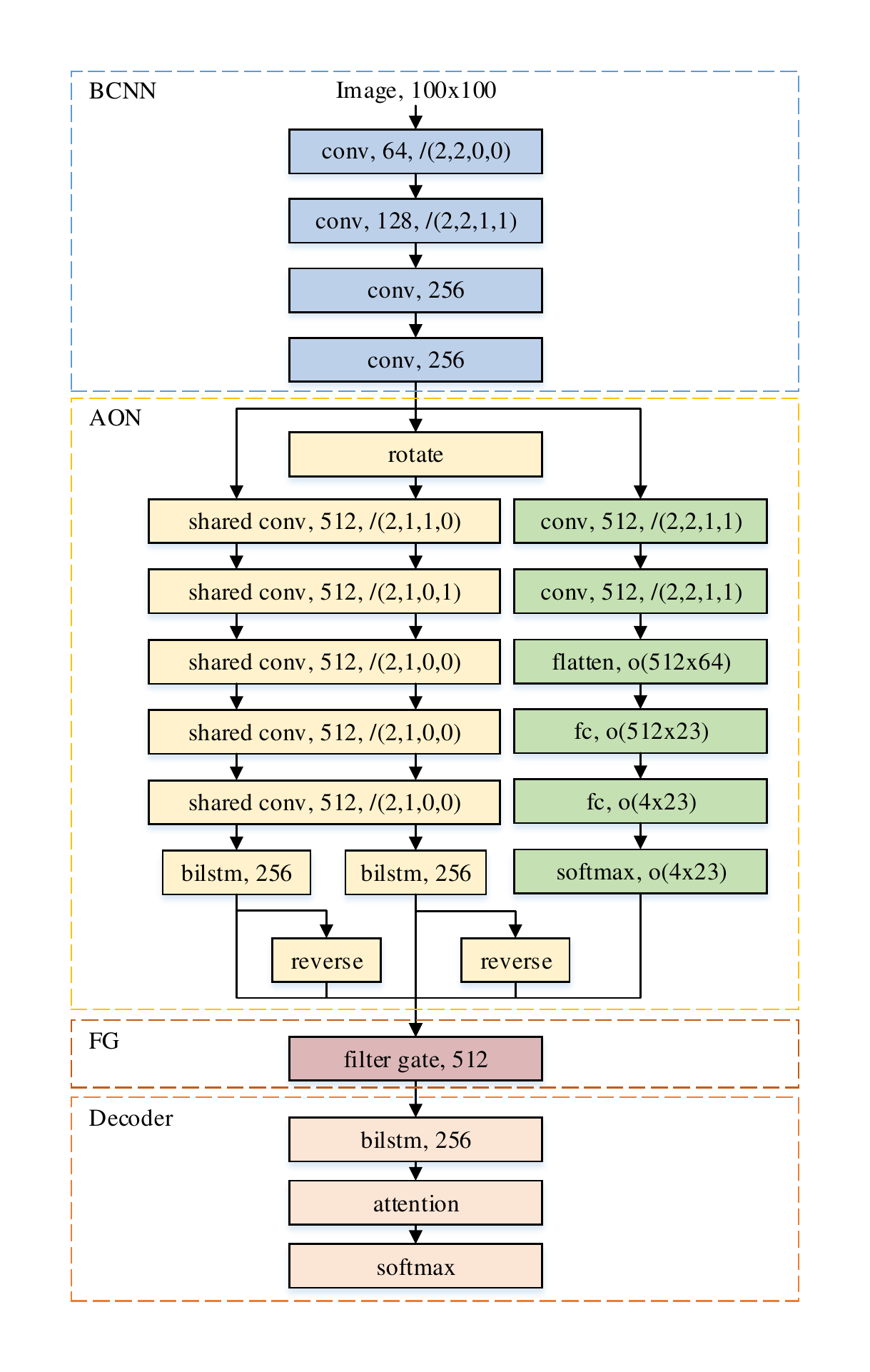}
    \end{center}
    \vspace{-2em}
    \caption{
    The network architecture of our method, which consists of four components:
        1) the basal convolutional neural network (BCNN) module for low-level visual representation;
        2) the arbitrary orientation network (AON) for capturing the horizontal, vertical and character placement features;
        3) the filter gate (FG) for combing four feature sequences with the character placement clues;
        4) the attention-based decoder (Decoder) for predicting character sequence. The above four modules are shown in the blue, golden, dull-red and brown dashed boxes, respectively.
        Meanwhile, all convolution or shared convolution blocks have the following format: $name, c[,/(s_h, s_w, p_h, p_w)]$. The bilstm and filter gate blocks are represented as $name, c$. The flatten, fc (fully-connected) and softmax operations have the format: $name, o(c, l)$. Here, $c$, $s_h$, $s_w$, $p_h$, $p_w$, $l$, $/$ and $o$ represent the number of channels, stride height, stride width, pad height, pad width, length of feature maps, pooling operation and output shape, respectively.
        The whole network can be trained end-to-end.
    }
    \label{fig:framework}
\end{figure}

The other approaches work in a top-down style: directly predicting the entire text from the original image without detecting the characters.
\cite{jaderberg2016reading} conducted a 90k-class classification task with a CNN, in which each class represents an English word. Consequently, the model can not recognize out-of-vocabulary words.
Recent works solve this problem as a sequence recognition problem, where images and texts are separately encoded as patch sequences and character sequences, respectively.
\cite{sutskever2014sequence} extracted sequences of HOG features to represent images, and generated the character sequence with the recurrent neural network (RNN).
\cite{He2015reading} and \cite{shi2016end} proposed the end-to-end neural networks that combines CNN and RNN for visual feature representation, then the CTC~\cite{Graves2006} Loss was combined with the RNN outputs for calculating the conditional probability between the predicted and the target sequences.
\cite{lee2016recursive} used a recursive CNN to learn broader contextual information, and applied the attention-based decoder for sequence generation.
\cite{cheng2017focus} proposed a focus mechanism to eliminate the attention drift to improve the regular text recognition performance.
However, since a text image is encoded into a 1D-based sequence of features, these methods can not effectively handle the irregular texts such as the arbitrarily-oriented texts.
In order to recognize irregular texts, \cite{shi2016robust} applied the spatial transformer network (STN) \cite{Jaderberg2015} for text rectification, then recognized the rectified text images with the sequence recognition network.
\cite{Yang2017attention} introduced an auxiliary dense character detection task for encouraging the learning of visual representations with a fully convolutional network (FCN) \cite{Long2015fcn}.
In practice, training STN-based methods is extremely difficult without human-labeled geometric ground truth, especially for texts in complicated (e.g. curved, arbitrarily-oriented or perspective etc.) environments.
Besides, sophisticated tricks are also required. For example, to train the thin-plate-spline (TPS) \cite{Bookstein1989}-based STN \cite{shi2016robust}-based method, the initialization pattern should be given for the fiducial points.
Though \cite{Yang2017attention} can recognize characters in a 2D image, the method relies on the multi-task learning framework (including 3 task branches and 2 tunable super-parameters) and character-level bounding box annotations, which results in large amount of resource consuming. While obtaining better performance on irregular texts, it performs worse on regular texts.

Different from existing approaches, in this paper we first extract deep feature representations of images by using an arbitrary orientation network (AON), then use a filter gate (FG) to generate the integrated sequence of features, which are fed to an attention-based decoder for generating predicted sequences.
Furthermore, we can once for all train the whole network end-to-end with only word-level annotations.

Note that in the OCR field, natural text reading systems often consist of two steps: 1) detecting each word's location in natural images and 2) recognizing text from the cropped image. In general, robust detection is helpful in recognizing texts. Therefore, several methods \cite{jiang2017r2cnn,ma2017arbitrary,shi2017detecting,zhang2016multi} have been proposed for multi-oriented text detection. Though this work focuses on the recognition task, our AON-based method can directly recognize arbitrarily-oriented texts, which alleviates the pressure of text detection.

\section{The Framework} \label{sec:arch}
The framework of whole network is shown in Fig.~\ref{fig:framework}, which consists of four major components:
1) The \emph{basal convolutional neural network} (a nomenclature for initial layer, denoted by BCNN) for extracting low-level visual features; 2) The \emph{arbitrary orientation network} (AON) for generating four-direction sequences of features and the character placement clues; 3)
    The \emph{filter gate} (FG) for combining the four sequences of features with the learned placement clues to generate the integrated feature sequence, and 4) the \emph{attention-based decoder} for predicting character sequence.

\subsection{Basal Convolutional Neural Network (BCNN)}
The BCNN module is responsible for capturing the foundational visual representation of text images, and outputs a group of feature maps.
BCNN can help reduce the computational cost and graphic memory.
As shown in Fig. \ref{fig:framework}, we use four convolution blocks as the foundational feature extractor.
The outputs of BCNN must be  square feature maps.
We empirically found that higher-level feature representation as the initial state of AON can yield better performance.

\subsection{Multi-Direction Feature Extraction Module}
This module includes the arbitrary orientation network (AON) and the filter gate (FG), which constitute the core of the proposed method. With the extracted foundational features, we devise AON for capturing arbitrarily-oriented text features and the corresponding character placement clues.
We also design FG for integrating multi-direction features by using the character placement clues.
The details of AON and FG will be described in next section.

\subsection{Attention-based Decoder}
An attention-based decoder is a recurrent neural network~(RNN) that directly generates the target sequence $(y_1, ..., y_M)$ from an input feature sequence $(\hat{h}_1, ..., \hat{h}_L)$. 
Bahdanau et al.~\cite{bahdanau2014neural} first proposed the architecture of attention-based decoder.
At the $t\text{-}th$ step, the attention module generates an output $y_t$ as follows: 

\begin{equation}
y_t = softmax(W^Ts_t),
\label{eq:py}
\end{equation}
where $W^T$ is a learnable parameter, and $s_t$ is the RNN hidden state at time $t$, computed by

\begin{equation}
s_t = RNN(y_{t-1}, g_t, s_{t-1}),
\label{eq:rnn}
\end{equation}
where $g_t$ is the weighted sum of sequential feature vectors $\hat{\mathcal{H}}:(\hat{h}_1, ..., \hat{h}_L)$, that is,

\begin{equation}
g_t = \sum_{j=1}^L \alpha_{t,j}\hat{h}_j,
\end{equation}
where $\alpha_t \in \mathbb{R}^L$ is a vector of the \textit{attention weights}, also called \textit{alignment factors} \cite{bahdanau2014neural}. 
In the computation of \textit{attention weights}, $\alpha_t$ is often evaluated by scoring each element in $\hat{\mathcal{H}}$ separately and normalizing the scores as follows:
\begin{equation}
\alpha_t = Attend(s_{t-1}, \hat{\mathcal{H}}),
\label{eq:attend}
\end{equation}
where $Attend$ describes the attending process \cite{chorowski2015attention}.

Above, the $RNN$ function in Eq.~(\ref{eq:rnn}) represents an LSTM recurrent network.
Note that the decoder is capable of generating sequences of variable lengths. Following \cite{sutskever2014sequence}, a special end-of-sequence (EOS) token is added to the target set, so that the decoder completes the generation of characters when EOS is emitted.

\subsection{Network Training}
We integrate the BCNN, AON, FG and attention decoder into one network, as shown in Fig. \ref{fig:framework}.
Therefore, given an input image $\mathcal{I}$, the loss function of the network is as follows:
\begin{equation}\label{eq:loss_attention}
\mathcal{L} = -\sum_{t}ln P(\hat{y}_t | \mathcal{I}, \theta),
\end{equation}
where $\hat{y}_t$ is the ground truth of the $t\text{-}th$ character and $\theta$ is a vector that combines all the network parameters.

\subsection{Character Sequence Decoding}
Decoding is the final process to generate the predicted characters.
Following the decoding conventions, two processing modes are given: unconstrained (lexicon-free) mode and constrained mode.
We execute unconstrained text recognition by directly selecting the most probable character.
While in constrained text recognition, with respect to different types of lexicons (their sizes are denoted by ``50'', ``1k'' and ``full'' respectively),
we calculate the conditional probability distributions for all lexicon words, and take the one with the highest probability as the output result.

\section{Technical Details of AON and FG}
\subsection{Arbitrary Orientation Network (AON)}
We develop an \emph{arbitrary orientation network}  consisting of
    the \emph{horizontal network }(HN), the \emph{vertical network} (VN) and
    the \emph{character placement clue network} (CN)
    for extracting horizontal, vertical and placement features respectively.

The HN encodes the foundational feature maps into a sequence of horizontal feature vectors $\mathcal{H}\in \mathbb{R}^{L\times D}$ by 
first performing downsampling on height directly by 5 shared convolutional blocks (described bellow) with the corresponding pooling strategy (shown in Fig. \ref{fig:framework}) to 1, and using the bidirectional LSTM to further encode the feature sequence, then generating the reversed feature sequence by conducting reverse operation (described in Eq. (\ref{eq:h}) and (\ref{eq:v})), where $L$ and $D$ represent the length of $\hat{\mathcal{H}}$ and the channel number, respectively.
Symmetrically, VN first rotates the square feature maps by 90 degrees, then generates the vertical feature vectors $\mathcal{V}\in \mathbb{R}^{L\times D}$ with the same procedure as HN.
Here, \emph{reversion} can accelerate training convergence, thus indirectly impacts the training of CN.

Since we describe each character sequence in four directions: $left \rightarrow right$, $right \rightarrow left$, $top \rightarrow bottom$ and $bottom \rightarrow top$, $\mathcal{H}$ and $\mathcal{V}$ can be represented as follows:
\begin{equation}
\mathcal{H} = \left\{
\begin{aligned}
\overrightarrow{\mathcal{H}}:(h_1, ..., h_L)^T,~~~~~~~left \rightarrow right &\\
\overleftarrow{\mathcal{H}}:(h_L, ..., h_1)^T, ~~~~~~~right \rightarrow left&
\end{aligned}
\right.
\label{eq:h}
\end{equation}
\begin{equation}
\mathcal{V} = \left\{
\begin{aligned}
\overrightarrow{\mathcal{V}}:(v_1, ..., v_L)^T,~~~~~~~top \rightarrow bottom &\\
\overleftarrow{\mathcal{V}}:(v_L, ..., v_1)^T. ~~~~~~~bottom \rightarrow top&
\end{aligned}
\right.
\label{eq:v}
\end{equation}

For each text image, the CN outputs the corresponding character placement
clues $\mathcal{C}\in \mathbb{R}^{4\times L}$ as:
\begin{equation}
\mathcal{C} = (c_1, ..., c_L)^T.
\end{equation}
Here, for any $c_i \in \mathbb{R}^4$, we have $\sum_{j=1}^4c_{ij} = 1$, where $c_{ij}$ refers to the $j\text{-}th$ direction's weight.
The extraction process of clues is depicted as the green blocks in Fig.~\ref{fig:framework}.

\begin{figure}[!htbp]
    \begin{center}
    \includegraphics[width=0.3\textwidth]{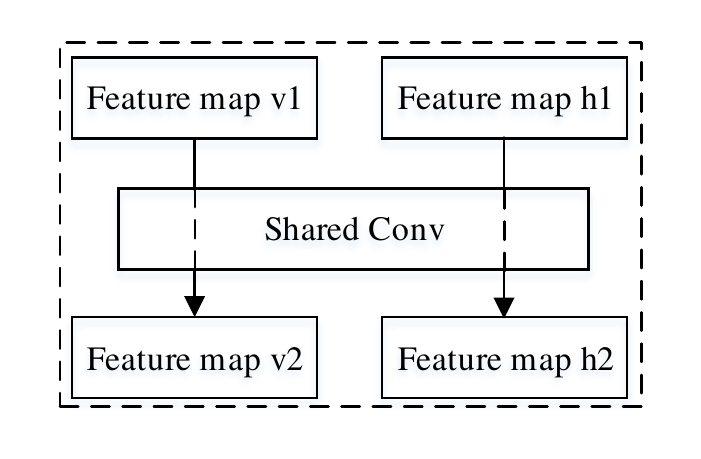}
    \end{center}
    \vspace{-1.5em}
    \caption{The shared convolution block in the dashed box provides a mechanism that multi-groups of feature maps share the same convolutional filters.
    }
    \label{fig:shareconv}
\end{figure}

In practice, we find that it is hard to train the HN and VN respectively and simultaneously.
The state of each branch is easy to corrupted on orientation distribution unbalanced training datasets. 
Therefore, we design a shared convolution mechanism that performs the same convolutional filter operations for both horizontal and vertical process, and the shared convolution block is shown in Fig. \ref{fig:shareconv}.
With the shared convolutional mechanism, the network is robust and easy to learn on orientation unbalanced training datasets.

\subsection{Filter Gate (FG)}
With the captured four feature sequences and character placement clues, we design a filter gate to neglect the irrelevant features. Formally, given the $i\text{-}th$ features $(\overrightarrow{\mathcal{H}}_i, \overleftarrow{\mathcal{H}}_i, \overrightarrow{\mathcal{V}}_i, \overleftarrow{\mathcal{V}}_i)$, we use the corresponding placement clue $c_i$ to attend the appropriate features:
\begin{equation}
 \hat{h}_i^\prime= \lbrack \overrightarrow{\mathcal{H}}_i~~\overleftarrow{\mathcal{H}}_i~~
 \overrightarrow{\mathcal{V}}_i~~\overleftarrow{\mathcal{V}}_i\rbrack  c_i.
\label{eq:py}
\end{equation}
Then an activation operation is performed as follows:
\begin{equation}
 \hat{h}_i= \tanh(\hat{h}_i^\prime).
\label{eq:py}
\end{equation}
Above, $\hat{h}_i$ indicates the $i\text{-}th$ element of $\hat{\mathcal{H}}: (\hat{h}_1, ..., \hat{h}_L)$.

\section{Performance Evaluation}\label{sec:experiments}
We conduct extensive experiments to validate the proposed method on both irregular and regular recognition benchmarks.
To be fair, we train the \emph{HN} used in AON as the baseline model (denoted by Naive\_base), which is similar to the previous works focusing on regular text recognition.
We also combine \emph{HN} with the TPS-based STN used in \cite{shi2016robust} as the STN-based control model (denoted by STN\_base). All control experiments are conducted with similar training data and in similar running environment.
We compare our model with not only the major existing methods (including the-state-of-the-art ones), but also the above two baseline models: Naive\_base and STN\_base.
Furthermore, we explore the roles of HN, VN, CN and FG in AON.
\subsection{Datasets}
The regular and irregular benchmarks are as follows:

{\bf{SVT-Perspective}} \cite{quy2013recognizing} contains 639 cropped images for testing. Images are picked from side-view angle snapshots in Google Street View, therefore one may observe severe perspective distortions. Each image is associated with a 50-word lexicon and a full lexicon.

{\bf{CUTE80}} (CT80 in short) \cite{risnumawan2014robust} is collected for evaluating curved text recognition. It contains 288 cropped natural images for testing. No lexicon is associated.

{\bf{ICDAR 2015}}~(IC15 in short) \cite{karatzas2015icdar} contains 2077 cropped images where more than 200 irregular (arbitrarily-oriented, perspective or curved). No lexicon is associated.

{\bf{IIIT5K-Words}} (IIIT5K in short) \cite{mishra2012scene} is collected from the Internet, containing 3000 cropped word images in its test set. Each image specifies a 50-word lexicon and a 1k-word lexicon, both of which contain the ground truth words as well as other randomly picked words.

{\bf{Street View Text}} (SVT in short) \cite{wang2011end} is collected from the Google Street View, consists of 647 word images in its test set. Many images are severely corrupted by noise and blur, or have very low resolutions. Each image is associated with a 50-word lexicon.

{\bf{ICDAR 2003}} (IC03 in short) \cite{lucas2003icdar} contains 251 scene images, labeled with text bounding boxes. Each image is associated with a 50-word lexicon defined by Wang et al. \cite{wang2011end}. For fair comparison, we discard images that contain non-alphanumeric characters or have less than three characters, following \cite{wang2011end}. The resulting dataset contains 867 cropped images. The lexicons include the 50-word lexicons and the full lexicon that combines all lexicon words.
\subsection{Implementation Details}\label{sec:implementation}
\emph{Network details:}
    The deep neural network has been detailed in Fig.~\ref{fig:framework}.
    In our network, all images are resized to $100 \times 100$.
    As for the convolutional strategy, all convolutional blocks have $3\times 3$ size of kernels, $1\times 1$ size of pads and $1\times 1$ size of strides, and all pooling (max) blocks have $2\times 2$ size of kernels.
    We adopt batch normalization (BN) \cite{ioffe2015batch} and ReLU activation right after each convolution.
    For the character generation task, the attention is designed with an LSTM~(256 memory blocks) and 37 output units (26 letters, 10 digits, and 1 EOS symbol).

\emph{Implementation and Running Environment:}
    We train our model on 8-million synthetic data released by Jaderberg et al.~\cite{jaderberg2014synthetic} and 4-million synthetic instances (excluding the images that contain non-alphanumeric characters) cropped from 80-thousand images \cite{Gupta16} by the ADADELTA \cite{adadelta} optimization method.
    Meanwhile, we conduct data augmentation by randomly rotating each image range from $0^\circ$ to $360^\circ$ once.
    Our method is implemented under the Caffe framework \cite{jia2014caffe}.
    The CUDA 8.0 and CUDNN v7 backend are extensively used in our implementation, so that most modules in our method are GPU-accelerated. Our method can handle about 190/630 samples per second in the training/testing phase.
    The experiments are carried out on a workstation with one Intel Xeon(R) E5-2650 2.30GHz CPU, one NVIDIA Tesla P40 GPU, and 128GB RAM.

\subsection{Performance on Irregular Datasets}
\begin{table}[!htbp]
  \begin{center}
    \begin{tabular}{|l|p{0.5cm}p{0.5cm}p{0.7cm}|p{0.7cm}|p{0.7cm}|}
    \hline
    \multirow{2}{*}{{Method}} &
    \multicolumn{3}{c|}{{SVT-Perspective}} & {CT80} & {IC15} \cr\cline{2-6}
      & {50} & {Full} & {None} & {None} & {None} \cr\hline
      ABBYY{\cite{wang2011end}}                          & 40.5    & 26.1 & $-$ & $-$  & $-$\cr
      Mishra \etal{\cite{graves2013speech}}         & 45.7 & 24.7 & $-$ & $-$   & $-$\cr
      Wang \etal{\cite{wang2012end}}                      &40.2  & 32.4  & $-$ & $-$  & $-$\cr
      Phan \etal{\cite{quy2013recognizing}}           & 75.6  & 67.0  & $-$ & $-$  & $-$\cr
       Shi \etal{\cite{shi2016end}}                    &92.6 & 72.6 & 66.8& 54.9  & $-$\cr
       Shi \etal{\cite{shi2016robust}}                    &91.2 & 77.4 & 71.8& 59.2  & $-$\cr
       Yang \etal{\cite{Yang2017attention}}        &93.0 & 80.2 & {\bf{75.8}}& 69.3  & $-$\cr
       Cheng \etal{\cite{cheng2017focus}}        &$92.6$ & $81.6$ & $71.5$& $63.9$  & 66.2\cr\hline
       Naive\_base & {{92.4}} & {{83.3}} & 70.5& {{75.4}}& {{67.8}} \cr
       STN\_base & {\bf{94.6}} & {{82.8}} & 68.5& {{73.7}}& {{67.5}} \cr
      Ours & {{94.0}} & {\bf{83.7}} & 73.0& {\bf{76.8}}& {\bf{68.2}} \cr\hline
    \end{tabular}
    \end{center}
  \caption{Results on irregular benchmarks. ``50'' is lexicon size and ``Full'' indicates the combined lexicon of all images in the benchmarks. ``None'' means lexicon-free.}
  \label{tab:results-irregular}
\end{table}
Recently, Cheng \etal \cite{cheng2017focus} proposed FAN to improve text recognition performance, which must be trained with additional character-level bounding box annotations. Here, we also compare our method with FAN on the irregular datasets.
Tab.~\ref{tab:results-irregular} summarizes the recognition results on three irregular text datasets: SVT-Perspective, CUTE80 and ICDAR15.
Comparing with the existing methods' performance results released in the literature,
we find that our method outperforms the existing methods on almost all benchmarks, except for SVT-Perspective with lexicon-free released by Yang \etal \cite{Yang2017attention}.
However, it is worthy of pointing out that Yang's method~\cite{Yang2017attention} implicates its text-reading system with both word-level and character-level bounding box annotations, which is resource consuming, while our method can be easily carried out with only word-level annotations.
\begin{figure}[!htb]
    \begin{center}
    \includegraphics[width=0.45\textwidth]{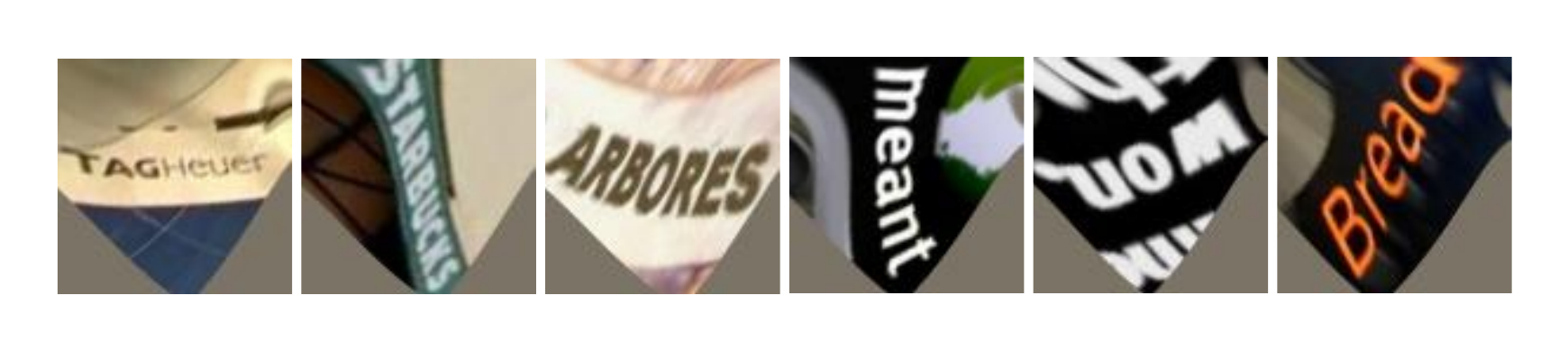}
    \end{center}
    \vspace{-1.2em}
    \caption{Some images rectified by TPS. }
    \label{fig:tps}
\end{figure}

Tab.~\ref{tab:results-irregular} also gives the results of the two baseline models.
We can see that Naive\_base does not recognize irregular texts well.
Though theoretically TPS-based STN can handle any irregular texts, it seems not able to satisfactorily rectify arbitrary-oriented or seriously curved texts in practice. Fig.~\ref{fig:tps} shows some rectified examples by TPS, their original images are shown in Fig.~\ref{fig:motivation}. We can see that except for the first and the third images, the other four images are not desirably rectified.
\begin{table}[!htbp]
  \begin{center}
  \scalebox{0.78}{
    \begin{tabular}{|l|p{0.4cm}p{0.4cm}p{0.6cm}|p{0.4cm}p{0.6cm}|p{0.4cm}p{0.4cm}p{0.6cm}|}
    \hline
    \multirow{2}{*}{{Method}} &
    \multicolumn{3}{c|}{{IIIT5k}} & \multicolumn{2}{c|}{{SVT}} & \multicolumn{3}{c|}{{IC03}}   \cr\cline{2-9}
      & {50} & {1k} & {None} & {50} & {None} & {50} & {Full} & {None}   \cr\hline
      ABBYY{\cite{wang2011end}} & 24.3          & $-$ & $-$ & 35.0 & $-$ & 56.0 & 55.0 & $-$   \cr
      Wang \etal{~\cite{wang2011end}}           & $-$  & $-$  & $-$ & 57.0 & $-$ & 76.0 & 62.0 & $-$   \cr
      Mishra \etal{\cite{graves2013speech}}         & 64.1 & 57.5 & $-$ & 73.2 & $-$ & 81.8 & 67.8 & $-$   \cr
      Wang \etal{\cite{wang2012end}}           &$-$  & $-$  & $-$ & 70.0 & $-$ & 90.0 & 84.0 & $-$   \cr
      Goel \etal{\cite{goel2013whole}}           &$-$  & $-$  & $-$ & 77.3& $-$ & 89.7& $-$  & $-$   \cr
      Bissacco \etal{\cite{bissacco2013photoocr}}       &$-$  & $-$  & $-$ & 90.4& 78.0& $-$ & $-$ & $-$  \cr
      Alsharif {\cite{alsharif2013end}}   &$-$  & $-$  & $-$ & 74.3& $-$ & 93.1 & 88.6 & $-$  \cr 
      Almaz{\'a}n \etal{\cite{almazan2014word}}         &91.2  & 82.1 & $-$ & 89.2 & $-$ & $-$ & $-$ & $-$   \cr
      Yao \etal{\cite{yao2014strokelets}}            &80.2 & 69.3 & $-$ & 75.9& $-$ & 88.5 & 80.3 & $-$   \cr
      Jaderberg \etal{\cite{jaderberg2014deep}}      &$-$  & $-$  & $-$ & 86.1& $-$ & 96.2& 91.5 & $-$   \cr
      Su and Lu{\cite{su2014accurate}}             &$-$  & $-$  & $-$ & 83.0& $-$ & 92.0& 82.0 & $-$   \cr
      Gordo{\cite{gordo2015supervised}}                 &93.3 & 86.6 & $-$ & 91.8& $-$ & $-$ & $-$ & $-$  \cr
      Jaderberg \etal{\cite{jaderberg2016reading}}      &97.1 & 92.7 & $-$ & 95.4& 80.7& {{98.7}}& {\bf{98.6}}& {{93.1}}  \cr
      Jaderberg \etal{\cite{jaderberg2014deep}}      &95.5 & 89.6 & $-$ & 93.2& 71.7& 97.8& 97.0& 89.6 \cr
      Shi \etal{\cite{shi2016end}}            &97.6 & 94.4 & 78.2& 96.4& 80.8& {{98.7}}& 97.6& 89.4 \cr
      Shi \etal{\cite{shi2016robust}}            &96.2 & 93.8 & 81.9& 95.5& 81.9& 98.3& 96.2&  90.1 \cr
      Lee \etal{\cite{lee2016recursive}}            &96.8 & 94.4 & 78.4& 96.3& 80.7& 97.9& 97.0&  88.7 \cr
      Yang \etal{\cite{Yang2017attention}}            &97.8 & 96.1 & $-$& 95.2& $-$& $-$& {{97.7}}&  $-$ \cr
      Cheng's baseline{\cite{cheng2017focus}}        &98.9 & 96.8 & {{83.7}}& {{95.7}}& {{82.2}}& {{98.5}}& 96.7&  {{91.5}} \cr
      Cheng \etal{\cite{cheng2017focus}}        &99.3 & 97.5 & {\bf{87.4}}& {\bf{97.1}}& {\bf{85.9}}& {\bf{99.2}}& 97.3&  {\bf{94.2}} \cr\hline
      Naive\_base & {{99.5}} & {\bf{98.1}} & 86.0& 96.9& 81.9& 98.5& 96.5 & 90.5   \cr
      STN\_base & {{99.5}} & {{97.8}} & 85.9& 96.3& 80.7& 98.5& 96.2 & 89.2   \cr
      Ours & {\bf{99.6}} & {\bf{98.1}} & 87.0& 96.0& 82.8& 98.5& 97.1 & 91.5   \cr\hline
    \end{tabular}
    }
    \end{center}
  \caption{Results on regular benchmarks. ``50'' and ``1k'' are lexicon sizes. ``Full'' indicates the combined lexicon of all images in the benchmarks. ``None'' means lexicon-free.}
  \label{tab:results-regular}
\end{table}

\subsection{Performance on Regular Datasets}
AON is designed for recognizing both irregular and regular texts.
Therefore, we test our method on some regular text benchmarks, the results are shown in Tab.~\ref{tab:results-regular}.
In the constrained cases, our method achieves comparable performance to the existing methods.
In the unconstrained cases, our method only falls behind Cheng \etal \cite{cheng2017focus} on the three benchmarks, and Jaderberg \etal \cite{jaderberg2016reading} on IC03.
For \cite{cheng2017focus}, two major factors lead to its high performance: a) using extra geometric annotations (location of each character) in training the attention decoder,
and b) exploiting a ResNet-based feature extractor for obtaining robust feature representation. However, labelling the location of each character is extremely expensive, so it is not feasible for real applications. For fair comparison, we also gave the results of Cheng's baseline (without the FocusNet branch) in Tab.~\ref{tab:results-regular}, and found that our method outperforms Cheng's baseline in most cases, which validates the superiority of our method.
Though Jaderberg \etal \cite{jaderberg2016reading} achieves an amazing results on IC03, their model cannot recognize out-of-vocabulary words, 
which limits its applicability in real world.
Note that our model is trained without any character geometric information, and it performs better than the other existing methods.
As a whole, our method performs effectively in recognizing regular texts.

\subsection{Deep insight into AON} 
Here, to further clarify the working mechanism of AON, we elaborate the roles of the major components \emph{HN}, \emph{VN},  \emph{CN} and \emph{FG} in AON, and show the placement trends of texts in some real images. These trends are generated by AON.
\begin{figure}[!htb]
    \begin{center}
    \includegraphics[width=0.45\textwidth]{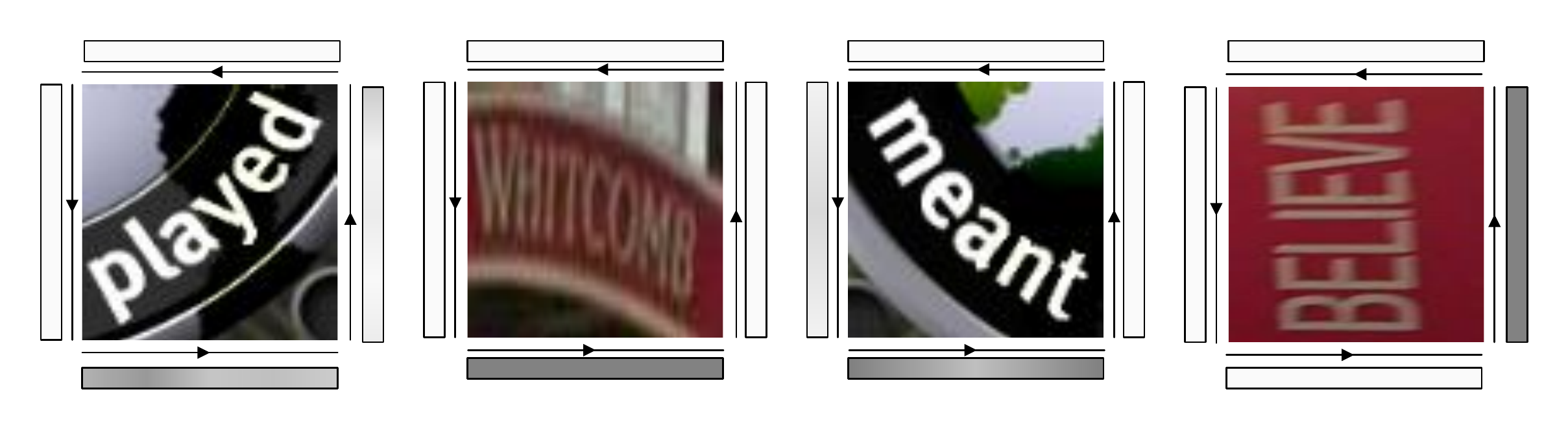}
    \end{center}
    \vspace{-1.2em}
    \caption{Illustration of learned character placement clues by AON. Each image is surrounded with four changing-gray bars with arrows of different directions. Deeper gray in the bars indicates larger weight for the corresponding directional feature.}
    \label{fig:bar}
\end{figure}

\textbf{The roles of \emph{HN},\emph{VN}, \emph{CN} and \emph{FG} in AON}.
We use both horizontal sequence of features and vertical sequence of features to represent arbitrarily-oriented texts.
Concretely, for horizontal/vertical texts, horizontal/vertical features are enough to represent the texts; For perspective/slant or arbitrarily-oriented text, we generate the final feature sequence by combining horizontal and vertical features.
\emph{HN} and \emph{VN} are responsible for generating horizontal and vertical features respectively.
\emph{CN} plays an important role in learning the weights (\ie, character placement clues) that are used to guide the generation of final feature sequences. 
\emph{FG} is just to perform the weight-sum operation with the horizontal/vertial feature sequence and the learnt placement weights.
Fig. \ref{fig:bar} shows some examples of generated placement clues. We can see that the generated clues conform to our visual observations in the images, which validates the effectiveness of \emph{CN} in AON.


\textbf{Text placement trends generated with AON}.
Here we verify that the learned character placement clues can be used to generate placement trends of character sequences by positioning each character and drawing text orientations in the original images. Bellow is the computation process of text placement trends.

We know that the  alignment factors $\alpha_t$ produced by the attention module indicate the probability distributions over the input sequence of features for generating the glimpse vector $g_t$.
And the four character placement clues $\mathcal{C}= [c_1, c_2, c_3, c_4]$ imply the importance of four extracted feature sequences for representing characters.
With $\mathcal{C}$ and $\alpha_t$, we roughly divide the input image into $L \times L$ patches and calculate the character position distribution $dis$ by $dis = \mathcal{C} \odot \alpha_t$,
where ${dis}=(d_1, d_2,d_3,d_4)\in \mathbb{R}^{4\times L}$.
We further normalize each element by 
$norm(d_{ij}) = \frac{d_{ij}}{\sum_{i=1}^2\sum_{j=1}^L d_{ij}}$
for $i$$\in$ (1, 2), and by
$norm(d_{ij}) = \frac{d_{ij}}{\sum_{i=3}^4\sum_{j=1}^L d_{ij}}$
for $i\in$ (3, 4). Here, $norm$ indicates the normalization operation.

For a character at position $(x, y)$, we first compute the horizontal coordinate $x$ with $[d_1, d_2]$ by
$x = \sum_{j=1}^L \sum_{i=1}^{2} j\times norm(d_{ij})$,
where $i\in (1, 2)$ and $j \in (1, 2, ..., L)$.
Similarity, we compute the vertical coordinate $y$ with $[d_3, d_4]$  by
$y = \sum_{j=1}^L \sum_{i=3}^{4} j\times norm(d_{ij})$.

\begin{figure}[!htbp]
    \begin{center}
    \includegraphics[width=0.45\textwidth]{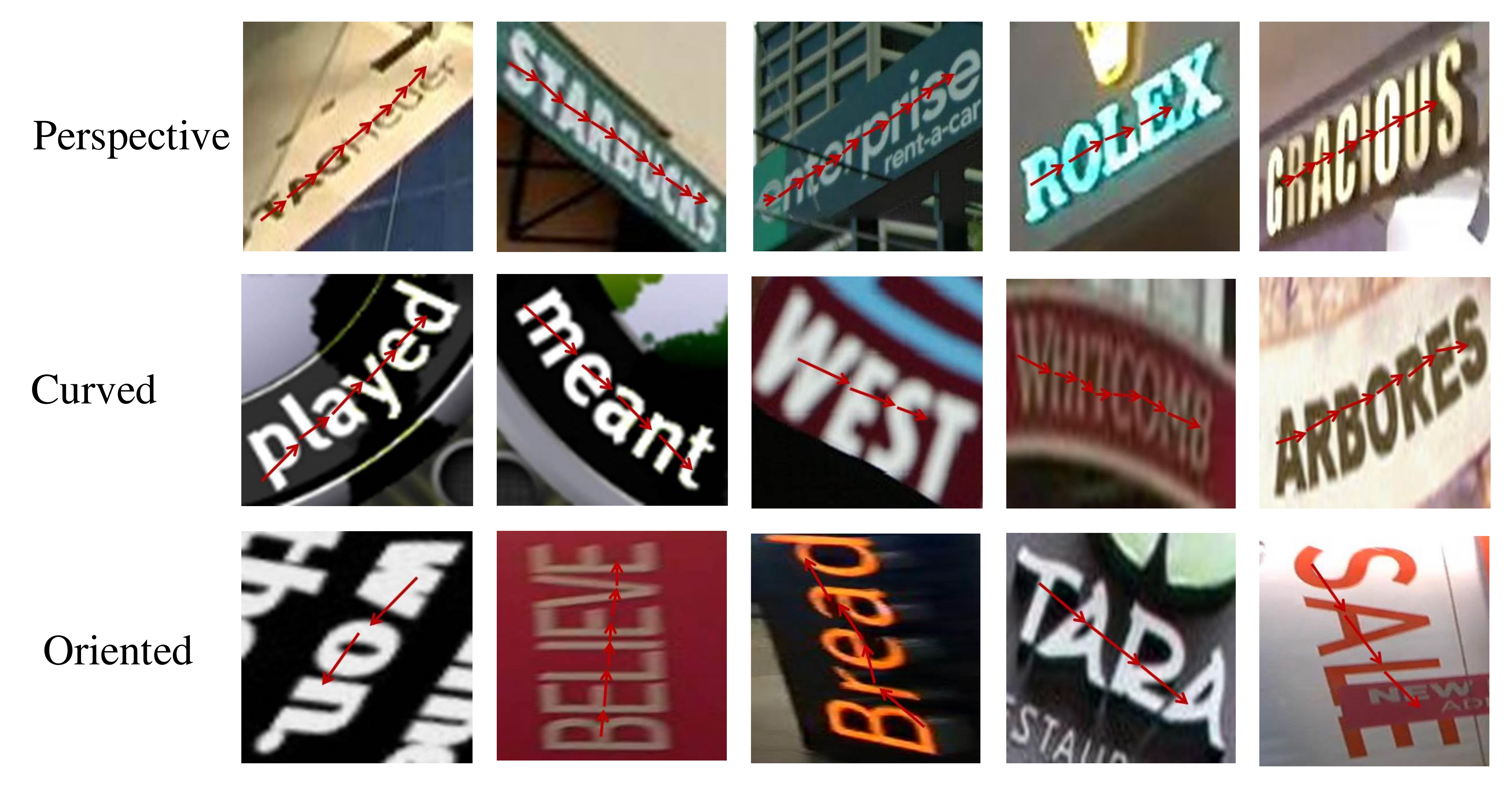}
    \end{center}
    \vspace{-1em}
    \caption{
    The visualization of generated placement trends for perspective, curved and oriented images, shown in the 1st, 2nd and 3rd row, respectively.
    The curves formed by connected red arrows indicate text placement trends.
    All texts in the images are correctly recognized by our method.
    }
    \label{fig:visual}
\end{figure}

To visualize the placement trends of texts in the input images, we mark the coordinate $(x, y)$ on each input image as the corresponding  character's position, and consecutively connect the last character's position and the current character's position with an arrow to describe the text's placement trend.
Fig.~\ref{fig:visual} shows some examples of generated text placement trends of real images. We can see that the trends formed by the connected arrows basically conform to our visual observations, which again shows that our method is effective in estimating the orientations of texts in images.
\section{Discussions}
\textbf{The necessity of CN in AON}.
As the attention-based decoder is able to select features for generating characters. It is natural to suspect whether \emph{CN} is necessary. To answer this, we have two experiments without \emph{CN}:
1) Concatenating horizontal and vertical feature sequence along the channel axis. We find the model converges slowly and cannot achieve state-of-the-art performance, because three quarters' information in the final feature sequence is superfluous.
2) Concatenating horizontal and vertical along the temporal axis. We get results of averagely about 4\% lower than that of AON on all benchmarks. The above experiments show that CN is important in AON.

\textbf{Impact of aspect ratio}. We studied the impact of aspect ratio by experiments, but did not observed obvious negative impact for images with a large aspect ratio. Compared to the previous works~\cite{shi2016end,shi2016robust}, the enlarging/shrinking operation in height does not obviously affect recognition results of horizontal texts with a large aspect ratio.

\textbf{Integrating with only two directional feature sequence}.
It is not reasonable to integrate only two directional sequences of features. For example, as shown in Fig.~\ref{fig:motivation}, if we integrate the right-left and down-top directional sequences of features to generate the final feature sequence, the visual features of `p' and `d' will be frame-wisely mixed up due to the weighting mechanism of FG.

\textbf{The computational cost of AON}.
Computational cost and the number of parameters are major concerns in resource-constrained scenarios such as embedded computer systems.
Comparing to the \emph{Naive\_base} model, the introducing of AON increases parameters and computational cost (twice of \emph{Naive\_base}).
However, the \emph{STN\_base} needs triple parameters and computational cost of \emph{Naive\_base}.
\section{Conclusion}
In this work, we propose a novel method to recognize arbitrarily oriented texts by 1) devising an arbitrary orientation network to extract visual features of characters in four directions and the character placement clues, 2) using a filter gate mechanism to combine the four-direction sequences of features, and 3) employing an attention-based decoder for generating character sequence.
Different from most existing methods, our method can effectively recognize both irregular and regular texts from images.
Experiments over both regular and irregular benchmarks validate the superiority of the proposed method.
In the future, we plan to extend the proposed idea to other related tasks.

\section{Acknowledgement} Fan Bai and Shuigeng Zhou were partially supported by the Program of Science and Technology Innovation Action of Science and Technology Commission
of Shanghai Municipality (STCSM) under grant No.~17511105204.

{\small
\bibliographystyle{ieee}
\bibliography{egbib}

\begin{thebibliography}{10}\itemsep=-1pt

\bibitem{almazan2014word}
J.~Almaz{\'a}n, A.~Gordo, A.~Forn{\'e}s, and E.~Valveny.
\newblock Word {S}potting and {R}ecognition with {E}mbedded {A}ttributes.
\newblock {\em IEEE TPAMI}, 36(12):2552--2566, 2014.

\bibitem{alsharif2013end}
O.~Alsharif and J.~Pineau.
\newblock End-to-end text recognition with hybrid hmm maxout models.
\newblock In {\em ICLR}, 2014.

\bibitem{bahdanau2014neural}
D.~Bahdanau, K.~Cho, and Y.~Bengio.
\newblock Neural {M}achine {T}ranslation by {J}ointly {L}earning to {A}lign and
  {T}ranslate.
\newblock In {\em ICLR}, 2015.

\bibitem{bissacco2013photoocr}
A.~Bissacco, M.~Cummins, Y.~Netzer, and H.~Neven.
\newblock Photo{OCR}: {R}eading {T}ext in {U}ncontrolled {C}onditions.
\newblock In {\em ICCV}, pages 785--792, 2013.

\bibitem{Bookstein1989}
F.~L. Bookstein.
\newblock {Principal warps: Thin-plate splines and the decomposition of
  deformations}.
\newblock {\em IEEE TPAMI}, 11(6):567--585, 1989.

\bibitem{cheng2017focus}
Z.~Cheng, F.~Bai, Y.~Xu, G.~Zheng, S.~Pu, and S.~Zhou.
\newblock {Focusing Attention: Towards Accurate Text Recognition in Natural
  Images}.
\newblock In {\em ICCV}, pages 5076--5084, 2017.

\bibitem{chorowski2015attention}
J.~K. Chorowski, D.~Bahdanau, D.~Serdyuk, K.~Cho, and Y.~Bengio.
\newblock Attention-{B}ased {M}odels for {S}peech {R}ecognition.
\newblock In {\em NIPS}, pages 577--585, 2015.

\bibitem{goel2013whole}
V.~Goel, A.~Mishra, K.~Alahari, and C.~V. Jawahar.
\newblock Whole is {G}reater than {S}um of {P}arts: {R}ecognizing {S}cene
  {T}ext {W}ords.
\newblock In {\em ICDAR}, pages 398--402, 2013.

\bibitem{gordo2015supervised}
A.~Gordo.
\newblock Supervised mid-level features for word image representation.
\newblock In {\em CVPR}, pages 2956--2964, 2015.

\bibitem{Graves2006}
A.~Graves, S.~Fern{\'a}ndez, F.~Gomez, and J.~Schmidhuber.
\newblock {Connectionist Temporal Classification : Labelling Unsegmented
  Sequence Data with Recurrent Neural Networks}.
\newblock In {\em ICML}, pages 369--376. ACM, 2006.

\bibitem{graves2013speech}
A.~Graves, A.~r.~Mohamed, and G.~Hinton.
\newblock Speech recognition with deep recurrent neural networks.
\newblock In {\em ICASSP}, pages 6645--6649, 2013.

\bibitem{Gupta16}
A.~Gupta, A.~Vedaldi, and A.~Zisserman.
\newblock Synthetic {Data for Text Localisation in Natural Images}.
\newblock In {\em CVPR}, pages 2315--2324, 2016.

\bibitem{He2015reading}
P.~He, W.~Huang, Y.~Qiao, C.~C. Loy, and X.~Tang.
\newblock {Reading Scene Text in Deep Convolutional Sequences}.
\newblock In {\em AAAI}, pages 3501--3508, 2016.

\bibitem{ioffe2015batch}
S.~Ioffe and C.~Szegedy.
\newblock Batch normalization: {A}ccelerating deep network training by reducing
  internal covariate shift.
\newblock In {\em ICML}, pages 448--456, 2015.

\bibitem{jaderberg2014synthetic}
M.~Jaderberg, K.~Simonyan, A.~Vedaldi, and A.~Zisserman.
\newblock Synthetic {D}ata and {A}rtificial {N}eural {N}etworks for {N}atural
  {S}cene {T}ext {R}ecognition.
\newblock {\em arXiv preprint arXiv:1406.2227}, 2014.

\bibitem{jaderberg2014deep}
M.~Jaderberg, K.~Simonyan, A.~Vedaldi, and A.~Zisserman.
\newblock Deep {S}tructured {O}utput {L}earning for {U}nconstrained {T}ext
  {R}ecognition.
\newblock In {\em ICLR}, 2015.

\bibitem{jaderberg2016reading}
M.~Jaderberg, K.~Simonyan, A.~Vedaldi, and A.~Zisserman.
\newblock Reading {T}ext in the {W}ild with {C}onvolutional {N}eural
  {N}etworks.
\newblock {\em IJCV}, 116(1):1--20, 2016.

\bibitem{Jaderberg2015}
M.~Jaderberg, K.~Simonyan, A.~Zisserman, and K.~Kavukcuoglu.
\newblock {Spatial Transformer Networks}.
\newblock {\em NIPS}, pages 2017--2025, 2015.

\bibitem{jia2014caffe}
Y.~Jia, E.~Shelhamer, J.~Donahue, S.~Karayev, J.~Long, R.~Girshick,
  S.~Guadarrama, and T.~Darrell.
\newblock Caffe: {C}onvolutional {A}rchitecture for {F}ast {F}eature
  {E}mbedding.
\newblock In {\em ACM-MM}, pages 675--678, 2014.

\bibitem{jiang2017r2cnn}
Y.~Jiang, X.~Zhu, X.~Wang, S.~Yang, W.~Li, H.~Wang, P.~Fu, and Z.~Luo.
\newblock R{2CNN: Rotational Region CNN for Orientation Robust Scene Text
  D}etection.
\newblock {\em arXiv preprint arXiv:1706.09579}, 2017.

\bibitem{karatzas2015icdar}
D.~Karatzas, L.~Gomez-Bigorda, A.~Nicolaou, S.~Ghosh, A.~Bagdanov, M.~Iwamura,
  J.~Matas, L.~Neumann, V.~R. Chandrasekhar, S.~Lu, F.~Shafait, S.~Uchida, and
  E.~Valveny.
\newblock I{CDAR} 2015 competition on {R}obust {R}eading.
\newblock In {\em ICDAR}, pages 1156--1160, 2015.

\bibitem{lee2016recursive}
C.~Y. Lee and S.~Osindero.
\newblock Recursive {R}ecurrent {N}ets with {A}ttention {M}odeling for {OCR} in
  the {W}ild.
\newblock In {\em CVPR}, pages 2231--2239, 2016.

\bibitem{Long2015fcn}
J.~Long, E.~Shelhamer, and T.~Darrell.
\newblock Fully convolutional networks for semantic segmentation.
\newblock In {\em CVPR}, pages 3431--3440, 2015.

\bibitem{lucas2003icdar}
S.~M. Lucas, A.~Panaretos, L.~Sosa, A.~Tang, S.~Wong, and R.~Young.
\newblock I{CDAR} 2003 robust reading competitions.
\newblock In {\em ICDAR}, pages 682--687, 2003.

\bibitem{ma2017arbitrary}
J.~Ma, W.~Shao, H.~Ye, L.~Wang, H.~Wang, Y.~Zheng, and X.~Xue.
\newblock A{rbitrary-Oriented Scene Text Detection via Rotation P}roposals.
\newblock {\em arXiv preprint arXiv:1703.01086}, 2017.

\bibitem{mishra2012scene}
A.~Mishra, K.~Alahari, and C.~V. Jawahar.
\newblock Scene {T}ext {R}ecognition using {H}igher {O}rder {L}anguage
  {P}riors.
\newblock In {\em BMVC}, pages 1--11, 2012.

\bibitem{neumann2012real}
L.~Neumann and J.~Matas.
\newblock Real-time scene text localization and recognition.
\newblock In {\em CVPR}, pages 3538--3545, 2012.

\bibitem{quy2013recognizing}
T.~Quy~Phan, P.~Shivakumara, S.~Tian, and C.~Lim~Tan.
\newblock Recognizing text with perspective distortion in natural scenes.
\newblock In {\em ICCV}, pages 569--576, 2013.

\bibitem{risnumawan2014robust}
A.~Risnumawan, P.~Shivakumara, C.~S. Chan, and C.~L. Tan.
\newblock A robust arbitrary text detection system for natural scene images.
\newblock {\em Expert Systems with Applications}, 41(18):8027--8048, 2014.

\bibitem{shi2017detecting}
B.~Shi, X.~Bai, and S.~Belongie.
\newblock D{etecting Oriented Text in Natural Images by Linking S}egments.
\newblock {\em arXiv preprint arXiv:1703.06520}, 2017.

\bibitem{shi2016end}
B.~Shi, X.~Bai, and C.~Yao.
\newblock An {End-to-End Trainable Neural Network for Image-based Sequence
  Recognition and Its Application to Scene Text Recognition}.
\newblock {\em IEEE TPAMI}, preprint, 2016.

\bibitem{shi2016robust}
B.~Shi, X.~Wang, P.~Lyu, C.~Yao, and X.~Bai.
\newblock Robust {Scene Text Recognition with Automatic Rectification}.
\newblock In {\em CVPR}, pages 4168--4176, 2016.

\bibitem{su2014accurate}
B.~Su and S.~Lu.
\newblock Accurate {Scene Text Recognition Based on Recurrent Neural Network}.
\newblock In {\em ACCV}, pages 35--48, 2015.

\bibitem{sutskever2014sequence}
I.~Sutskever, O.~Vinyals, and Q.~V. Le.
\newblock Sequence to {Sequence Learning with Neural N}etworks.
\newblock In {\em NIPS}, pages 3104--3112, 2014.

\bibitem{wang2011end}
K.~Wang, B.~Babenko, and S.~Belongie.
\newblock End-to-end scene text recognition.
\newblock In {\em ICCV}, pages 1457--1464, 2011.

\bibitem{wang2010word}
K.~Wang and S.~Belongie.
\newblock Word {Spotting in the W}ild.
\newblock In {\em ECCV}, pages 591--604. Springer, 2010.

\bibitem{wang2012end}
T.~Wang, D.~J. Wu, A.~Coates, and A.~Y. Ng.
\newblock End-to-end text recognition with convolutional neural networks.
\newblock In {\em ICPR}, pages 3304--3308, 2012.

\bibitem{xu2016show}
K.~Xu, J.~Ba, R.~Kiros, K.~Cho, A.~Courville, R.~Salakhudinov, R.~Zemel, and
  Y.~Bengio.
\newblock {Show, Attend and Tell: Neural Image Caption Generation with Visual
  Attention}.
\newblock In {\em ICML}, pages 2048--2057, 2015.

\bibitem{Yang2017attention}
X.~Yang, D.~He, Z.~Zhou, D.~Kifer, and C.~L. Giles.
\newblock {Learning to Read Irregular Text with Attention Mechanisms}.
\newblock In {\em IJCAI}, pages 3280--3286, 2017.

\bibitem{yao2014strokelets}
C.~Yao, X.~Bai, B.~Shi, and W.~Liu.
\newblock Strokelets: {A} {L}earned {M}ulti-scale {R}epresentation for {S}cene
  {T}ext {R}ecognition.
\newblock In {\em CVPR}, pages 4042--4049, 2014.

\bibitem{ye2015text}
Q.~Ye and D.~Doermann.
\newblock Text {D}etection and {R}ecognition in {I}magery: {A} {S}urvey.
\newblock {\em IEEE TPAMI}, 37(7):1480--1500, 2015.

\bibitem{adadelta}
M.~D. Zeiler.
\newblock {ADADELTA:} {A}n {A}daptive {L}earning {R}ate {M}ethod.
\newblock {\em CoRR}, abs/1212.5701, 2012.

\bibitem{zhang2016multi}
Z.~Zhang, C.~Zhang, W.~Shen, C.~Yao, W.~Liu, and X.~Bai.
\newblock Multi-oriented text detection with fully convolutional networks.
\newblock In {\em CVPR}, pages 4159--4167, 2016.

\end{thebibliography}
}

\end{document}